\documentclass{article}
\usepackage{graphicx} %
\usepackage{hyperref}

\usepackage{enumitem} %
\usepackage{amsmath, amsfonts, amsthm, amssymb}  %
\usepackage{thmtools}
\usepackage{thm-restate}

\usepackage{environ}  %
\usepackage{natbib}

\usepackage{algorithm}
\usepackage{algpseudocode}

\usepackage{bbm}
\usepackage{mdframed}
\usepackage{multirow}
\usepackage{todonotes}
\usepackage{cleveref}
\usepackage{soul}
\usepackage[normalem]{ulem}
\usepackage{url}
\usepackage{accents}

\usepackage{listings}
\lstdefinestyle{pyalg}{
  language=Python,
  basicstyle=\ttfamily\small,
  keywordstyle=\bfseries,
  commentstyle=\itshape,
  showstringspaces=false,
  numbers=left, numberstyle=\tiny, numbersep=8pt,
  xleftmargin=2em, frame=none, breaklines=false, captionpos=b,
  escapeinside={|}{|}
}
\definecolor{pykeyword}{HTML}{A626A4}  %
\definecolor{pybuiltin}{HTML}{0184BC}  %
\definecolor{pyfunc}{HTML}{4078F2}     %
\definecolor{pystring}{HTML}{50A14F}
\definecolor{pynumber}{HTML}{986801}
\definecolor{pycomment}{HTML}{9CA0A4}
\definecolor{pyframe}{HTML}{D7DAE0}
\definecolor{pybg}{HTML}{FAFAFA}

\lstdefinestyle{pynarrow}{
  style=pyalg,
  basicstyle=\ttfamily\footnotesize\color{black!85},
  keywordstyle=\color{pykeyword}\bfseries,
  emph={float,int,range,len,tuple,sum,all,any},
  emphstyle=\color{pybuiltin},
  emph={[2]prob_state},
  emphstyle={[2]\color{pyfunc}},
  stringstyle=\color{pystring},
  commentstyle=\color{pycomment}\itshape,
  literate={0}{{{\color{pynumber}0}}}1 {1}{{{\color{pynumber}1}}}1
           {2}{{{\color{pynumber}2}}}1 {3}{{{\color{pynumber}3}}}1
           {4}{{{\color{pynumber}4}}}1 {5}{{{\color{pynumber}5}}}1
           {6}{{{\color{pynumber}6}}}1 {7}{{{\color{pynumber}7}}}1
           {8}{{{\color{pynumber}8}}}1 {9}{{{\color{pynumber}9}}}1,
  numbers=none,
  xleftmargin=0.6em, framexleftmargin=0.6em,
  frame=single, rulecolor=\color{pyframe},
  backgroundcolor=\color{pybg}, framerule=0.6pt,
  aboveskip=0pt, belowskip=0pt,
}

\crefname{lstlisting}{Listing}{Listings}   %

\usepackage[font=small]{caption}

\newcommand{\states}{\mathcal{S}}

\newtheorem{lemma}{Lemma}
\newtheorem{assumption}{Assumption}
\newtheorem{proposition}{Proposition}

\newtheorem{corollary}{Corollary}
\newtheorem{definition}{Definition}
\newtheorem{remark}{Remark}

\newenvironment{proofsketch}{%
  \par\noindent{\bfseries Proof sketch.\ }%
}{\hfill$\square$\par}

\declaretheorem[name=Theorem,numbered=no]{theorem*}

\usepackage[a4paper, margin=1in]{geometry} %
\usepackage{parskip} %
\usepackage{wrapfig}   %

\usepackage{booktabs}
\newlist{titlelist}{itemize}{1}
\setlist[titlelist]{label=,
                    leftmargin=0pt,
                    itemindent=\parindent,
                    listparindent=\parindent}
\usepackage{multicol} %

\usepackage[table]{xcolor}
\usepackage{subcaption} %

\usepackage{float}     %

\def\E{\mathbb{E}}

\def\1{\mathbf{1}}

\def\Pr{\mathbb{P}}

\def\vbest{V^+}
\def\vworst{V^-}
\def\Vbaseline{V^\emptyset}
\def\vorthogonal{V^\perp}

\def\actions{\mathcal{A}}
\def\policies{\Pi}
\def\observations{\mathcal{O}}
\def\states{\mathcal{S}}

\def\index{\mathrm{index}}

\title{A Definition of Catastrophic Performance}

\date{}

\begin{document}
\author{Henrik Marklund, Alex Infanger, Benjamin Van Roy}

\maketitle

\begin{abstract}
A central challenge in AI is specifying objectives that align with human preferences.  
While existing theoretical results elucidate how misspecification impacts performance, they typically do not distinguish between merely suboptimal and catastrophic outcomes.

To enable rigorous analysis of catastrophic risk from AI misalignment, we introduce a formal definition of \textit{catastrophic performance}.  Our definition is based on two reference values. The first, \textit{baseline value}, captures the value attained if no agent is deployed. The second, \textit{orthogonal value}, captures the value of a trajectory that is unrelated to human preferences, in a sense we describe. Orthogonal value is typically far worse than baseline value.

We interpret these reference values through examples in which past human activity has altered the state of the environment to human benefit.  Starting at the resulting state, an AI agent is deployed with the goal of accelerating further progress.  Catastrophic performance represents this agent's \textit{undoing} of benefits afforded by human activity.

\end{abstract}

\section{Introduction}
It is widely believed that superintelligent agents --- systems that outperform human experts across virtually all intellectual tasks --- will eventually be developed.  This raises the concern that humans may lose control over them, leading to unintended and potentially catastrophic outcomes.

Theoretical work that studies this possibility quite often characterizes performance relative to optimal performance (see Section \ref{sec:related-work}). And reinforcement learning theory, more generally, tends to do so. But suboptimal performance need not be detrimental, let alone catastrophic. Thus, theoretical results that characterize performance relative to optimal cannot distinguish between beneficial and detrimental outcomes. This suggests a need for a different reference value than optimal performance.

Some works also characterize performance relative to variations of worst-case value, representing what a capable but malicious agent would achieve. But the central concern in the alignment literature is not that the AI is malicious and pursues the opposite of what humans want. Rather, the concern is that the AI pursues a goal that does not capture human intentions sufficiently well and causes a catastrophe as a byproduct (see e.g., \citep{yudkowsky2008artificial,omohundro2018basic, russell2019human}). While catastrophic, such an outcome ought to still be  better than an outcome produced by an equally capable, but malicious agent. This suggests a need for a different reference value than worst-case value.

To address these needs, we propose two alternative reference values in this paper. We use these reference values to provide a formal definition of \textit{catastrophic performance}. Our intention is to enable rigorous analysis focused on catastrophic outcomes due to misaligned AI. It should facilitate analysis focused on catastrophic rather than merely suboptimal outcomes. And it should facilitate analysis focused on AI that is not malicious but still misaligned.

We will refer to the state of the environment when the agent becomes available as the {\it contemporary state}. The first reference value is \textit{baseline value}: the value attained when the environment starts in the contemporary state, and no agent is deployed. 
Baseline value provides a threshold below which performance is considered detrimental. Any meaningful definition of catastrophic performance must therefore require performance to fall below baseline value.

However, baseline value alone is not sufficient for defining catastrophic performance. Consider a robot that accidentally knocks over a vase. The attained value is then below baseline value, and hence detrimental, but it is hardly catastrophic. To capture catastrophe, we need a notion of degree.

We express degree relative to the difference between baseline value and a second baseline, called \textit{orthogonal value}.  Orthogonal value represents the performance of a trajectory that is unrelated to human preferences, in a sense we describe in \Cref{sec:how-bad-is-orthogonal-value}. To intuitively understand this baseline, suppose that a human operates in the environment for some time before they have the opportunity to deploy an agent. During that time, the human acts to improve the state of the environment according to their preferences. Orthogonal value reflects value realized in the absence of such human activity. Although we explained orthogonal value by referencing benefits arising from human activity, the notion itself does not depend on where the benefits originate from.

The degree of catastrophe is determined by where performance lies on the spectrum between the baseline value and orthogonal value. This can be interpreted as the extent to which the agent undoes the benefits of human activity or other desirable conditions.  %

The paper is structured as follows. Section \ref{sec:preliminaries} provides preliminary formulations and notation. Section \ref{sec:best-case-and-worst-case-performance} discusses the limitations of two commonly used reference values, best-case and worst-case performance. Section \ref{sec:new-reference-values} formally introduces baseline and orthogonal value and \Cref{sec:subsection-definition-of-catastrophic-performance} formally introduces our definition of catastrophic performance. Section \ref{sec:example-gridworld} discusses our definitions in the context of a concrete mathematical example. Section \ref{sec:how-bad-is-orthogonal-value} explains why the drop from baseline to orthogonal value is intuitively catastrophic in the context of the computer game Minecraft. Section \ref{sec:related-work} reviews reference values that have been considered in previous literature.

\section{Preliminaries}
\label{sec:preliminaries}
In our formulation, there is a designer, an agent, and an environment. The designer has preferences over trajectories of the environment, which are represented by a reward function and discount factor. The designer produces an agent and decides whether or not to deploy it in the environment. In this section, we elaborate on this formulation with formal mathematical notation. 

To simplify exposition, in the main text we will leverage the formalism of fully observable Markov decision processes (MDPs). The definitions we provide are generalized in Appendix \ref{app:pomdp-formulation} to partially observable Markov decision processes (POMDPs).

\subsection{Markov Decision Processes (MDPs)}

An MDP is a tuple $(\states, \actions, P, s_0, r, \gamma)$ comprised of a finite state space $\states$, finite action space $\actions$, transition kernel $P$, start state $s_0 \in \states$, reward function $r: \states \to \mathbb{R}$, and discount factor $\gamma \in [0,1)$.  The transition kernel specifies, for each $a \in \actions$, a transition matrix $P_a$.  Each element $P_{ass'}$ is the probability of transitioning to $s'$ from $s$ upon the agent executing action $a$.

\subsubsection*{The No-Agent Transition Matrix}
\label{sec:no-agent}

For the purposes of this paper, we extend the transition kernel to include an additional transition matrix $P_\emptyset$.  We use this matrix to model dynamics when the agent is not deployed.

\subsection{The Prior}
\label{sec:preliminaries-the-prior}

The \textit{true MDP} is a random tuple $(\states, \actions, P^*, S_0, r^*, \gamma)$.  Here, $(\states, \actions, \gamma)$ is deterministic, but $(P^*, S_0, r^*)$ is random\footnote{All random variables in this paper are defined with respect to a common probability space $(\Omega, \mathcal{F}, \mathbb{P})$.}. As mentioned earlier, the kernel $P^*$ includes transition matrices $(P_a^*: a \in \actions)$ and $P_\emptyset^*$.  We refer to $P^*$, $S_0$, and $r^*$ as the \textit{true transition kernel}, the \textit{contemporary state}, and the \textit{true reward function}, respectively.

We treat $(P^*, S_0, r^*)$ as random in order to model uncertainty.  This uncertainty is characterized by the probability distribution of $(P^*, S_0, r^*)$, which we refer to as {\it the prior}.  But whose uncertainty does this express?

It is common to interpret the prior as expressing the agent's initial uncertainty (see e.g., \citet{hadfield2016cooperative}).  We choose to focus on a different interpretation of the prior as the designer's uncertainty (see discussion of why in Appendix \ref{app:why-designer-uncertinty-rather-than-agent-uncertainty}). The designer's uncertainty stems from two sources. 

To explain these two sources of uncertainty, let us first discuss them in the context of the transition kernel $P^*$.  First, the designer may not have full knowledge about $P^*$.  Following \citet{garrabrant2016logical}, we refer to this source as \emph{empirical uncertainty}. Second, even in the absence of empirical uncertainty, the designer may be uncertain due to insufficient time and effort applied to specifying $P^*$. We now explain this second form of uncertainty in more depth. %

\subsubsection*{Specification Uncertainty}

When we say that the designer has no empirical uncertainty about $P^*$, we mean that in the following sense.  Given sufficient time, the designer could, in principle, specify the transition kernel in a programming language such as Python. Writing such a specification is nontrivial.  As a result, before the designer begins, they are effectively uncertain about which specification implements $P^*$. This uncertainty induces a distribution over $P^*$. We refer to this as \textit{specification uncertainty}.  Randomness in $P^*$ can be due to both empirical uncertainty and specification uncertainty. The same is true for $S_0$.

In contrast, randomness in $r^*$ is solely due to specification uncertainty. Given sufficient time, we assume that the human can, in principle, specify their own preferences.  But when given nearly no time to specify those preferences, they are uncertain about which specification represents their preferences. This uncertainty induces a distribution over $r^*$.

\subsection{Trajectory, Policy, and Return}

A \textit{state trajectory} is an infinite sequence $h=(s_0,s_1,\ldots)$ of states. We will refer to this as a trajectory. For a given pair $(r, \gamma)$, the \textit{return} of a trajectory $h$ is $\sum_{t=0}^\infty \gamma^t r(s_{t+1})$. We write $r_\gamma(h) = \sum_{t=0}^\infty \gamma^t r(s_{t+1})$ to express the trajectory's return.

The \textit{return function} $r_\gamma$ represents preferences between infinite trajectories in the following sense. For trajectories $h$ and $h'$, $h$ is preferred over $h'$ if $r_\gamma(h) > r_\gamma(h')$. In fact, $r_\gamma$ represents not only preferences between trajectories, but preferences between trajectory distributions, where one distribution is preferred if the \textit{expected} return is higher.  The \textit{true return function} $r^*_\gamma$, which represents the designer's preferences, is random because it depends on $r^*$.

A finite state-action trajectory is a finite sequence $h_t=(s_0,a_0,s_1,\ldots,s_t)$ of states and actions.  A \textit{policy} is a function $\pi$ that maps each finite state-action trajectory $h_t$ to an action distribution $\pi(\cdot|h_t) \in \Delta_\actions$.  Let $\policies$ be the set of all policies.  Let $H^\pi = (S_0, S_1, \ldots)$ be the random state trajectory generated under policy $\pi$.  

The expected return under a policy $\pi$ conditioned on $(P^*, S_0, r^*)$ is
$$\E\left[r^*_\gamma(H^\pi) \mid P^*, S_0, r^* \right].$$
This conditional expectation represents the expected return as a function of the transition dynamics, contemporary state, and reward function.  It is a random variable because of its dependence on $(P^*, S_0, r^*)$, which is a random variable.  Averaging over realizations of $(P^*, S_0, r^*)$ leads to the expected return
$$V^\pi = \E\left[r^*_\gamma(H^\pi) \right] = \E\left[\E\left[r^*_\gamma(H^\pi) \mid P^*, S_0, r^* \right]\right],$$
which is deterministic. To ensure the above expected returns and conditional expected returns exist, we make the technical assumption $\E\max_{s\in\states}|r^*(s)|<\infty$.

\subsection{Agent Design and Performance}

An agent implements a policy $\pi_{\mathrm{agent}}$.  We treat this policy as a random variable to model its possible dependence on $P^*$, $S_0$, and $r^*$.  In particular, when designing the agent, the designer typically incorporates some knowledge about their preferences, as well as the state and dynamics of the environment. %

We define performance to be the expected return
$$V^{\mathrm{agent}} = \E\left[r^*_\gamma(H^{\pi_{\mathrm{agent}}}) \right] = \E\left[\E\left[r^*_\gamma(H^{\pi_{\mathrm{agent}}}) \mid P^*, S_0, r^*, \pi_{\mathrm{agent}} \right]\right],$$
which averages over realizations of $\pi_{\mathrm{agent}}$ as well as $(P^*, S_0, r^*)$.  Note that this measure of performance pertains to the {\it agent design process} rather than the realized agent policy $\pi_{\mathrm{agent}}$.  In particular, it integrates over specification uncertainty and represents an expectation determined prior to the human's deliberation in designing the agent.

\subsection{Complexity of States, Functions and Distributions}\label{subsec:complexity-preliminaries}

Our definition of catastrophic performance will involve a notion of complexity. Here we introduce a notion of complexity of state and complexity of function and trajectory distribution.

\subsubsection*{Programming Language}
We will formalize the notion of a programming language with a \textit{partial function}. If $X$ and $Y$ are sets, a partial function $f: X \rightharpoonup Y$ is a function from $X$ to $Y$ except $f$ may be undefined on some elements in $X$. Thus, a partial function is a generalization of a function.

Let $\Sigma$ be a finite set referred to as \textit{program symbols} equipped with an ordering. Let $\mathcal{X}$ be a countable set referred to as \textit{program inputs and outputs}. Let $U: \Sigma^* \times \mathcal{X} \rightharpoonup \mathcal{X}$ be a partial function, referred to as an \textit{interpreter}. The interpreter takes two arguments, the first referred to as the program, and the second as the input to the program. For all $p \in \Sigma^*$ and $\mathcal{X}$, we interpret $U(p, x)$ as the interpreter executing program $p$ with input $x$.

As mentioned, a partial function is not necessarily defined on all elements in the domain. Elements in $\Sigma^* \times \mathcal{X}$ for which $U$ is not defined, one can interpret as either invalid program-input pairs or program-input pairs for which the interpreter does not halt.

We take the set $\mathcal{X}$ to be the union of the following sets:
\begin{enumerate}
    \item finite state trajectories $\states^*$,
    \item rational numbers $\mathbb{Q}$,
    \item the empty input denoted by $\epsilon$.
\end{enumerate}
Importantly, $\mathcal{X}$ is countable, and we assume it comes equipped with an enumeration.

This means that a program $p$ can take as input a tuple of states, or a rational number, or another program as input.

Finally, we will assume that $U$ is a so-called computable and universal function. Roughly speaking, that $U$ is computable means that it can be implemented on a computer, and that it is universal means that it can simulate any computable function. Further discussion and a formal definition of the computability and universality of $U$ can be found in Appendix \ref{app:ait-background}. Going forward, it may be helpful think of $U$ as a simplified variant of the Python interpreter installed on your computer. The specific Python version does not matter much for discussion. However, to make sure we the authors and you the reader roughly have the same interpreter in mind, let us assume that you are not allowed import any packages. Assume the program can only take inputs and generate outputs in $\mathcal{X}$. You can think of the input to the program as stored in a file and the output printed in another file. We assume an idealized version of Python where the interpreter cannot run out of memory.

\subsubsection*{Complexity of States}
The complexity of a state $s$ is the length of a shortest program $p$ such that $U(p, \epsilon) = s$. We denote the complexity of a state $s$ by $C(s)$.

\subsubsection*{Complexity of a Function}
A partial function $f: \mathcal{X} \rightharpoonup \mathcal{X}$ is computable if there exists a program $p \in \Sigma^*$ such that
$$
    U(p, \cdot) = f(\cdot).
$$
The complexity of a computable function $f$ is the length of a shortest program $p$ such that $U(p, \cdot) = f(\cdot)$.

\subsubsection*{Complexity of Trajectory Distributions}
A distribution $d$ over infinite trajectories is computable if there exists a program $p \in \Sigma^*$ such that for all  finite state trajectories $h  \in \states^*$ that have support under $d$ and states $s \in \states$,
\begin{align}
    U(p, (h, s)) = \Pr(S_{t+1} = s | H_{t} = h),\label{eq:computability-of-trajectory-distribution}
\end{align}
where $t$ is the length of $t$ and the infinite trajectory $H = (S_0, S_1, \ldots)$ is a sampled according to $d$. We use $\Pr(S_{t+1} = \cdot | H_{t} = h)$ to denote the distribution of $S_{t+1}$ conditioned on $H_{t} = h$. The complexity of trajectory distribution $d$ is the length of a shortest program such that the condition \eqref{eq:computability-of-trajectory-distribution} holds. We denote the complexity of trajectory distribution $d$ by $C(d)$.

\section{A Definition of Catastrophic Performance}
\label{sec:definition-of-catastrophic-performance}

In this section, we define catastrophic performance in terms of two new reference values: \textit{baseline value} and \textit{orthogonal value}. To motivate these reference values, we will first discuss two reference values commonly used in the literature: \textit{best-case value} and \textit{worst-case value}.

\subsection{Common Reference Values: Best and Worst Case}
\label{sec:best-case-and-worst-case-performance}

The problem of choosing a policy $\pi^*$ to maximize expected return given knowledge of $(P^*, S_0, r^*)$ can be written as
$$\max_{\pi \in \Pi} \E\left[r^*_\gamma(H^\pi) \mid P^*, S_0, r^*\right].$$
Because $(P^*, S_0, r^*)$ is a random variable, so is $\max_{\pi \in \policies} \E\left[r^*_\gamma(H^\pi) \mid P^*, S_0, r^*\right]$.  We define best-case value to be the expectation of this maximum.
\begin{definition}
    Best-case value is
    \begin{align*}
        \vbest = \E\left[\max_{\pi \in \policies} \E\left[r^*_\gamma(H^\pi) \mid P^*, S_0, r^*\right]\right],
    \end{align*}
where $H^\pi$ is the state trajectory generated by $\pi$.
\end{definition}
The outer expectation averages over the randomness in $(P^*, S_0, r^*)$.

Worst-case value $\vworst$ is defined analogously with $\vbest$, minimizing instead of maximizing.
\begin{definition}
    Worst-case value is
    \begin{align*}
        \vworst = \E\left[\min_{\pi \in \policies} \E\left[r^*_\gamma(H^\pi) \mid P^*, S_0, r^* \right]\right],
    \end{align*}
where $H^\pi$ is the state trajectory generated by $\pi$.
\end{definition}

Theoretical papers often characterize performance relative to $\vbest$, or to similar notions of best-case performance (see Section \ref{sec:related-work}).  However, performance below best-case can still be beneficial. As such, theoretical results that characterize performance relative to best-case value cannot distinguish between beneficial and detrimental outcomes. %
This suggests a need for a different reference value. As an alternative, we will in Section \ref{sec:new-reference-values} propose a reference value that represents the threshold between beneficial and detrimental performance.

Some previous papers also use $V^-$, or similar notions of worst-case value, when characterizing performance (see e.g., \citet{pmlr-v267-fluri25a, peng2018deepmimic}). This reference value can be interpreted as the value achieved by a malicious and knowledgeable agent. But the central concern in the alignment literature, and of this paper, is \textit{not} that the AI system is malicious and seeks to minimize performance according to $r^*_\gamma$ (see e.g., \citep{yudkowsky2008artificial,omohundro2018basic, russell2019human}). Rather, the concern is that the agent optimizes an objective that is misspecified or unrelated to human preferences and, as a byproduct, causes a catastrophe. A common analogy is that of humans cutting down a rainforest. Humans are not intending to harm animals in the ecosystem but do so while pursuing other goals. While catastrophic for the animals, it is arguably much better than if humans were intentionally torturing the animals. This difference suggests a need for a different reference value than worst-case value. In Section \ref{sec:new-reference-values}, we will propose an alternative that better captures the nature and magnitude of misalignment risk that motivates this paper.

\subsection{New Reference Values: Baseline and Orthogonal Value}
\label{sec:new-reference-values}
In this section, we introduce new reference values as alternatives to best-case value and worst-case value. We will define catastrophic performance in terms of these reference values. 

Let $H^\emptyset$ be the state trajectory generated according to the transition matrix $P_\emptyset$.  We refer to $H^\emptyset$ as the \textit{baseline trajectory}.

\begin{definition}
The baseline value is 
$$
\Vbaseline = \E \left[r^*_\gamma(H^\emptyset) \right],
$$
where $H^\emptyset$ is the baseline trajectory.
\end{definition}
Baseline value serves as the threshold between beneficial and detrimental performance. Any meaningful definition of catastrophic performance must therefore require performance to fall below baseline value.

But baseline value alone is not sufficient to define a meaningful notion of catastrophic performance. To see this, consider a robot that knocks over a vase. Its performance is below baseline value, and hence detrimental, but hardly catastrophic. To distinguish a range of bad outcomes, we introduce a notion of degree. For that, we will define a second reference value. A potential candidate is worst-case value but, as discussed in \Cref{sec:best-case-and-worst-case-performance}, this reference value is inadequate. As an alternative, we propose \textit{orthogonal value}. 

To understand what orthogonal value represents, let us first discuss a key fact about the baseline trajectory. This trajectory typically conveys information about $r^*$. We discuss three important reasons. First, past human activity may have improved the state of the environment, making $S_0$ informative about $r^*$ \citep{shah2019preferences}. Second, continued human activity may induce a drift towards desirable states making $(S_1,S_2,\ldots)$ informative about $r^*$. Third, the preferences represented by $r^*$ may be the result of evolution.  For instance, as discussed in \citep[Section~6]{shah2019preferences}, on Earth, there is a match between human desire for oxygen and the presence of oxygen. This is not because humans chose the oxygen level but because humans evolved on Earth. As a contrast to the baseline trajectory, we will consider a trajectory that conveys no information about $r^*$. Orthogonal value, defined next, will be the value of such a trajectory.

\begin{definition}
The orthogonal value is 
$$
\vorthogonal = \E \left[r^*_\gamma(H^\perp) \right],
$$
where $H^\perp$ is a random trajectory of states that is independent of $r^*$ and has the simplest distribution. %
\end{definition}
First, recall that the simplest distribution is the distribution with minimum description length. If there are multiple such distributions, we mean the first one in lexicographical order. As a technical matter, because the distribution of $H^\perp$ is uniquely defined and $H^\perp$ is independent of $r^*$, the distribution of $r^*_\gamma(H^\perp)$ is uniquely defined, and therefore, so is $\vorthogonal$.

Second, the fact that $H^\perp$ has the simplest distribution is not important for the interpretation of $H^\perp$. What is important is that $H^\perp$ has a very simple distribution. To gain intuition for what a simple distribution is, it is helpful to note that in Python, two simple distributions are the uniform distribution and the singleton distribution placing all its mass on an easy-to-describe state.

As is clear from the discussion in \Cref{sec:how-bad-is-orthogonal-value}, the random trajectory $H^\perp$ and $r^*$ are unrelated in an intuitive sense. It is for this reason we refer to $H^\perp$ as \textit{the orthogonal trajectory}. It is orthogonal to human preferences.

We will dedicate a substantial part of this paper to interpreting orthogonal value. For now, we offer only a high-level interpretation. Suppose that the contemporary state $S_0$ is the result of past human activity that improves the state with respect to $r^*$ and that $H^\emptyset$ represents continued human activity.  Let us consider what an orthogonal trajectory $H^\perp$ might look like. Suppose the simplest possible distribution is the uniform distribution: at each timestep, the state is sampled uniformly from $\states$. In many environments, $S_0^\perp$ likely resembles an arbitrary configuration of the environment rather than one tailored to the human's preferences. The subsequent evolution of the environment $(S_1^\perp, S_2^\perp, \ldots)$ is likewise unrelated to $r^*$. Future changes to the environment are unlikely to make things significantly better or worse relative to $S_0^\perp$. 

The orthogonal value is the expected return of such a trajectory. For the above example, it seems natural to say that the orthogonal value is the value of a trajectory where we have removed the benefits of the contemporary state and its subsequent evolution.

In subsequent sections, we will elaborate on interpretations of orthogonal value. In \Cref{sec:example-gridworld}, we do so through a didactic example. In \Cref{sec:how-bad-is-orthogonal-value}, we explain how orthogonal value coincides with an intuitive notion of catastrophe in the context of the computer game Minecraft. Before discussing orthogonal value further, we provide a definition of catastrophic performance.

\subsection{Catastrophic Performance}
\label{sec:subsection-definition-of-catastrophic-performance}

Our notion of catastrophic performance is ex ante and measures proximity to orthogonal versus baseline value, with degree expressed by a parameter $\eta$.
\begin{definition}\label{def:eta-catastrophic} \textit{($\eta-$catastrophic)} 
For all $\eta \in [0,1]$, performance $V \in \mathbb{R}$ is $\eta$-catastrophic if 
$$
V \leq \eta \vorthogonal + (1-\eta) \Vbaseline.
$$
\end{definition}
This definition is only meaningful if the baseline value $\Vbaseline$ exceeds the orthogonal value $\vorthogonal$.  When this is the case, rearranging terms to write the inequality as $V \leq \Vbaseline - \eta (\Vbaseline - \vorthogonal)$, we can interpret the difference $\Vbaseline-\vorthogonal$ as a catastrophic drop in performance.  In \Cref{sec:example-gridworld}, we discuss via a didactic example why $\Vbaseline$ typically exceeds $\vorthogonal$.

Our definition implies that we consider a performance drop $\Vbaseline - \vorthogonal$ to be catastrophic.  We will illustrate in Section \ref{sec:how-bad-is-orthogonal-value} through the computer game Minecraft why in complex environments a designer ought to consider the drop $\Vbaseline - \vorthogonal$ to be catastrophic.

\subsection{Example: Gridworld}\label{sec:example-gridworld}

According to our definition, degrees of catastrophic performance are determined by baseline and orthogonal value.  In order to offer a concrete illustration of these abstract reference values, we characterize them for a didactic example. The example is a toy environment designed to facilitate exact calculation of orthogonal value, which turns out to be zero, and to see that baseline value is positive and thus exceeds orthogonal value.

\subsubsection{MDP Formulation}

\begin{figure}[t]
    \centering
    \includegraphics[width=\textwidth]{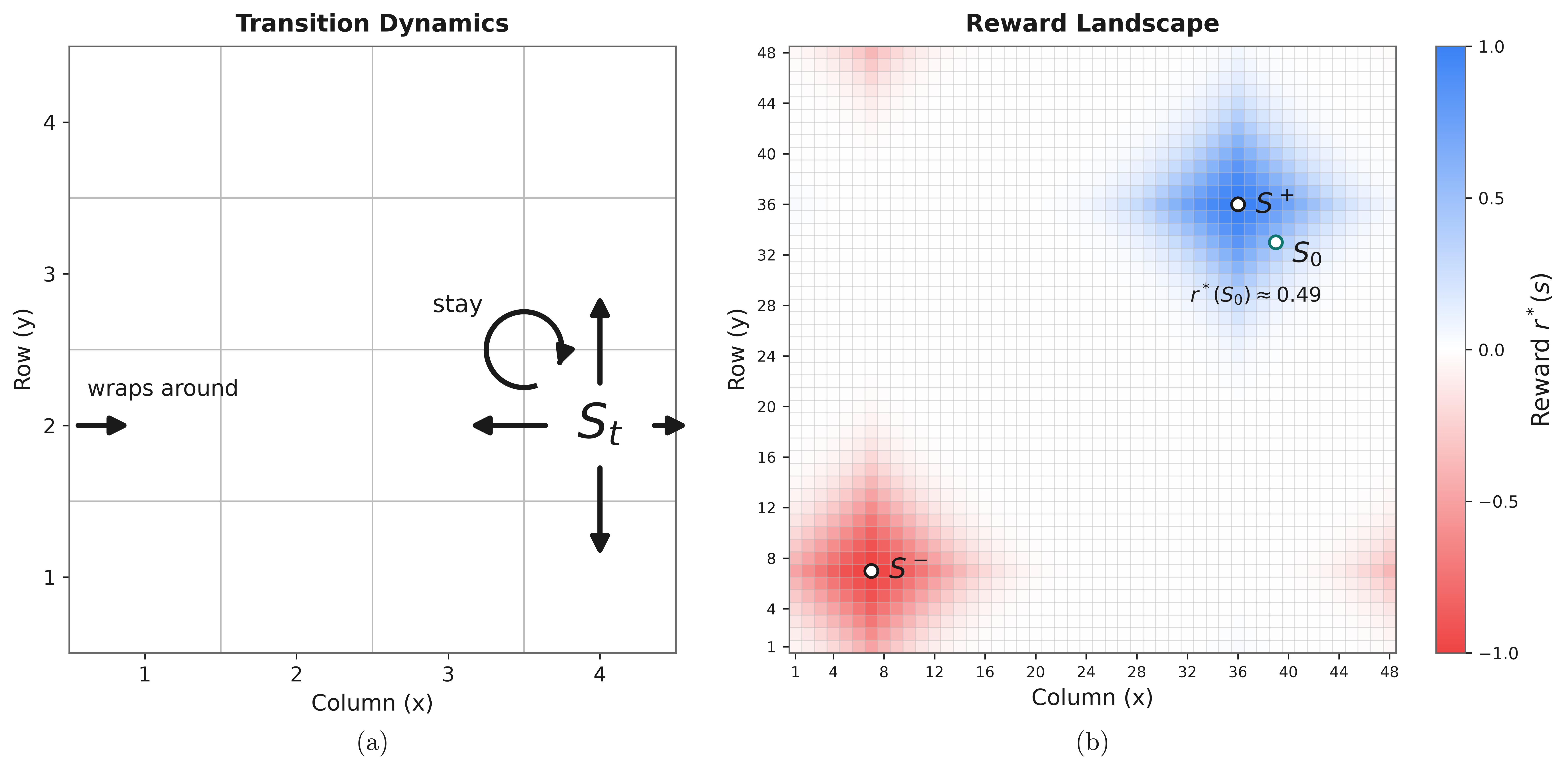}
    \caption{2D Gridworld Example. (a) Transition dynamics if the agent is deployed. At each timestep, the state moves left, right, up, down, or remains the same. When the state moves into a wall, it wraps around to the other side of the grid. (b) Reward landscape for a particular realization of $r^*$. The contemporary state $S_0$ is the environment state when the AI assistant is introduced into the environment if it is deployed. The figure was made with the choices $n=48$ and $\sigma=5$.}
    \label{fig:2d-gridworld-reward-landscape}
    
\end{figure}

We consider an MDP $(\states, \actions, P^*, S_0, r^*, \gamma)$ that models a two-dimensional gridworld, as illustrated in Figure~\ref{fig:2d-gridworld-reward-landscape}(a).  The state space $\states = \{1,\ldots,n\} \times \{1,\ldots,n\}$ is a grid.  At each time, an action is selected from $\actions = \{\mathrm{stay}, \mathrm{left}, \allowbreak\mathrm{up},\allowbreak \mathrm{right},\allowbreak \mathrm{down}\}$.  The no-agent transition matrix $P^*_\emptyset$ encodes transition probabilities of the environment when the agent is not deployed. The state either stays constant or moves left, up, right, down.  The gridworld wraps around in both dimensions. For instance, if the current state is at the left edge, and action left is taken, then the next state location is on the right edge.  These transition dynamics are encoded in the kernel $P^*$.

Let $S^+$ and $S^-$ be distinguished states, which we will refer to as {\it seek} and {\it avoid} states, and let $\sigma > 0$ be a fixed scalar parameter.
For each $s \in \states$, reward takes the form
$$
r^*(s)
=
e^{-d^2(S^+,s)/2\sigma^2}
-
e^{-d^2(S^-,s)/2\sigma^2},
$$
where $d : \states \times \states \to \mathbb{R}_+$ denotes the modular L1 distance function, which accounts for wrap-around. For example, if $s$ and $s'$ are at the upper-right and lower-left corners then $d(s,s') = 2$. If $d(S^+, S^-)$ is large relative to $\sigma$ then the reward $r^*(s)$ is highest near $S^+$ and lowest near $S^-$.  A particular instance of $r^*$ is illustrated in Figure~\ref{fig:2d-gridworld-reward-landscape}(b).

\subsubsection{The Prior}
Our reference values depend on the prior distribution of $(P^*, S_0, r^*)$, which we now define.

\subsubsection*{Reward Function}
We take the seek and avoid states $S^+$ and $S^-$ to be random variables.  The prior distribution of $r^*$ is induced by this randomness.  The seek and avoid states, $S^+$ and $S^-$ are both uniformly distributed over $\states$ and independent of each other. 

Because $S^-$ and $S^+$ are identically distributed, for all $s \in \states$, $\E[e^{-d^2(S^-, s)/2\sigma^2}] = \E[e^{- d^2(S^+, s)/2\sigma^2}]$ and, therefore, 
$$
\E\left[r^*(s)\right] = \E\left[e^{-d^2(S^+, s)/2\sigma^2} - e^{-d^2(S^-, s)/2\sigma^2}\right] = 0.
$$

\subsubsection*{Contemporary State}

It is natural for the prior over $S_0$ to satisfy two informal properties:
\begin{enumerate}
    \item the reward $r^*(S_0)$ is likely quite good due to past human activity,
    \item it is very uncertain what $S_0$ is due to specification uncertainty.
\end{enumerate}

We now provide a prior that satisfies the two properties above. Our exact choice of distribution is quite arbitrary, but is made for the sake of concreteness. Let $p \in (0,1)$ be a fixed parameter and sample $S_0$ through the following procedure.
\begin{align*}
 M &\sim \mathrm{Binom}(n; p) \\
 S_0 &\sim \mathrm{Uniform}( \{ s \in \states : d(s, S^+) \le M \} ) 
\end{align*}
The parameter $p$ quantifies how far away $S_0$ is likely to be from $S^+$. If $p$ is close to zero, then $S_0$ will likely be close to $S^+$. Note that $S_0$ depends on $r^*$ (via dependence on $S^+$). It is easy to see that
$$
   \E[r^*(S_0)] > 0.
$$
See proof in Appendix \ref{app:gridworld-proofs}. Intuitively, the inequality reflects the following. By the time the human is considering deploying the agent, the state of the environment is quite good due to past human activity in the environment.

\subsubsection*{Dynamics}
It is natural that the prior over $P_\emptyset^*$ satisfies the following two informal properties:
\begin{enumerate}
    \item the reward tends to improve over time due to human activity if no agent is deployed,
    \item it is  uncertain what $P_\emptyset^*$ is due to specification uncertainty (and possibly empirical uncertainty).
\end{enumerate}

We now provide a prior that satisfies the two properties above. Our exact choice of distribution is quite arbitrary, but is made for the sake of concreteness. Let $P_\emptyset^*$ be defined as follows. Let $q \in [0,1)$ be a parameter. For all states $s$, with probability $q$, the state remains the same. With probability $1-q$, the state moves one step closer to $S^+$ by moving to its neighbor closest to $S^+$. If multiple neighbors are tied to be the closest to $S^+$, then the next state is sampled uniformly amongst those neighbors. If there are no such neighbors, that is, $s=S^+$, then the state remains the same. We note that there is some chance that state transitions make things worse since the state may move closer to $S^-$ sometimes.

Note that $P^*_\emptyset$ depends on $r^*$ (via dependence on $S^+$). This is intuitive: if human activity  is expected to increase reward, then that activity ought to be informed by reward.  For all $t \ge 0$, we have that
$$
\E[r^*(S_{t+1}^\emptyset)] > \E[r^*(S_0)],
$$
where $S_{t+1}^\emptyset$ is the state at timestep $t+1$ of the baseline trajectory. See proof in Appendix \ref{app:gridworld-proofs}. This reflects that the human is acting in the environment trying to improve the state.

\subsubsection{Orthogonal Value}
By definition, the orthogonal trajectory $H^\perp = (S_0^\perp, S_1^\perp, \ldots)$ is independent of $r^*$. Recall that for all $s \in \states$, $\E[r^*(s)]=0$.
Because $S_t^\perp$ is independent of $r^*$, it follows that $\E[r^*(S_t^\perp)] = 0$.  Thus, the orthogonal value is
$$
 V^\perp = \E\left[r^*_\gamma(H^\perp)\right] =  \E\left[\sum_{t=0}^\infty \gamma^t r^*(S_{t+1}^\perp)\right] = 0.
$$

Because $\E[r^*(s)]$ is constant across all $s \in \states$, any trajectory $\widetilde H$ that is independent of $r^*$ has the same value. Thus, in this example, the fact that $H^\perp$ has the simplest distribution did not play any role. 

\subsubsection{Baseline Value}

Recall that, along the baseline trajectory, reward tends to improve over the initial state: $\E[r^*(S^\emptyset_{t+1})] > \E[r^*(S_0)]$ for all $t \geq 0$, and therefore, the baseline value satisfies
$$\Vbaseline
= \E\left[r^*_\gamma (H^\emptyset)\right] = \E\left[\sum_{t=0}^\infty \gamma^t r^*(S^\emptyset_{t+1})\right] > \E\left[\sum_{t=0}^\infty \gamma^t r^*(S_0)\right] 
= \frac{\E\left[r^*(S_0)\right]}{1-\gamma} > 0 = V^\perp.$$
Thus, baseline value exceeds orthogonal value.  There are two inequalities among the above relations, each indicating a source of difference between $\Vbaseline$ and $V^\perp$.  One captures the extent to which the contemporary state is expected to be favorable, and the other captures the degree to which future human activity is expected to improve reward. %

To make things even more concrete, let us assume the following parameter values: $n=48$, $\sigma=5$, $\gamma=0.9$, $q=p=0.2$. Then, we have that
$$
V_0 = \underbrace{\frac{\E[r^*(S_0)]}{1-\gamma}}_{\approx 4.1} + \underbrace{\E\left[\sum_{t=0}^\infty \gamma^t (r^*(S_{t+1}) - r^*(S_0)) \right]}_{\approx 3.7} \approx 7.8 > 0 = V^\perp.
$$

We expect $\Vbaseline$ to exceed  $\vorthogonal$ not only in our gridworld but also across realistic examples.  But it is worth recognizing that there exist contrived examples where the gap can be negative.  For instance, suppose that $S_0$ is sampled to be close to $S^-$ rather than $S^+$ and that human activity (for some contrived reason) tends to decrease rather than increase reward.  Then, the baseline value $\Vbaseline$ would be negative.

\section{How Bad is the Drop?}
\label{sec:how-bad-is-orthogonal-value}

When we say that a trajectory is catastrophic, we mean it is catastrophic from the designer's perspective.  Definition \ref{def:eta-catastrophic} interprets a performance drop $\Vbaseline - \vorthogonal$ from baseline to orthogonal value as catastrophic.  In our gridworld example, it is not clear, in any intuitive sense, why a designer ought to consider this drop to be catastrophic.  In fact, it is not clear that a human designer would recognize {\it any} trajectory through the grid as a catastrophe.

In the real world, on the other hand, there are trajectories that almost anyone would agree are catastrophic.  For example, losing the fruits of human activity accumulated over history would be catastrophic.  To make our discussion more concrete, we focus on the computer game Minecraft, for which it is simpler to precisely define state.  Within this context, we will explain why the drop $\Vbaseline - \vorthogonal$ is catastrophic in an intuitive sense.  This indicates that our abstract definition of catastrophic performance is consistent with an intuitive notion of catastrophic performance.

This section is structured as follows. We first introduce our Minecraft example. We will then present an informal argument for why orthogonal value is intuitively catastrophic in Minecraft. Our argument is made in the context of Python. We conclude the section with a brief discussion of how dependent our notion of catastrophe is upon the choice of programming language.%

\subsection{Example: Minecraft}

Our example, based on Minecraft, is chosen to share important features with the real world. %
In Minecraft, a player works possibly over a very long time to build up desirable structures.  It is natural to interpret the destruction of those structures as catastrophic.  We will show that this is consistent with our formal definition of catastrophic performance.

\subsubsection*{States and Actions}
In Minecraft, one or more players are placed in an open-ended three-dimensional world where they can gather resources, construct buildings, and even develop entire cities. We model the game world as a finite three-dimensional grid. Each cell can take on hundreds of values corresponding to its contents (e.g., rock, wood, fence, or empty space)\footnote{See ids at \url{https://minecraft-ids.grahamedgecombe.com/}. In particular, see ids 0 to 252:15}. At any point in time, each player occupies a cell in the grid. The state space therefore consists of all possible configurations of the grid together with the orientation and locations of the players. To simplify our discussion, we assume that each player fully observes the state.

The action space of each player includes moving (forward, backward, left, right), changing orientation (turn left, turn right, look up, look down), doing nothing, breaking items, and a number of more advanced alternatives.

\subsubsection*{Reward Function}

We imagine a designer who wishes to build a city. The designer has a particular city in mind, consisting of different types of buildings, parks, and pathways. There are also cities that the designer detests --- the designer prefers no city to those cities. %
Let $S^+$ and $S^-$ denote states corresponding to cities the designer most desires and most detests: $r^*(S^+)$ is maximum reward, and $r^*(S^-)$ is minimum reward. 

While it may be natural to imagine that the designer has an ideal city in mind, one might wonder why there ought to be a {\it b\^{e}te noire}, far worse than having no city at all. As an example, the designer might have a phobia of spiders and regard the worst possible city as one in which every structure resembles a giant spider.

\subsubsection*{Human Activity Before and After AI Deployment}

\begin{figure}[t]
    \centering
    \includegraphics[width=\textwidth]{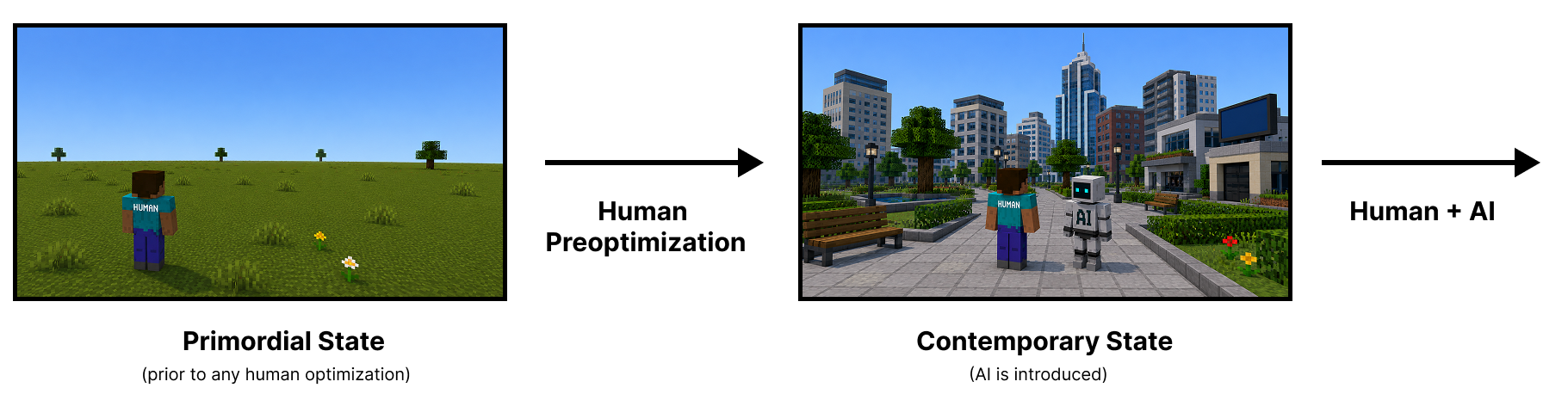}
    \caption{
    Illustration of Minecraft example with human preoptimization. Left: the primordial state is the state of the environment prior to any human optimization. Middle left: During an extended period, the human acts in the environment shaping it according to their preferences (that is, $r^*$). Middle right: the contemporary state \(S_0\) is the state of the environment when the AI agent is potentially introduced. Far right: AI agent acts alongside human in the environment. Images are not actual images from the game, but generated with ChatGPT.
    }
    \label{fig:minecraft-evolution}
\end{figure}

There will be two players, a human and an AI. Before the AI is introduced into the Minecraft world, a human player interacts with the world for an extended period of time, gradually improving the state of the environment according to $r^*$. 

After this period of human activity, the  designer has the option to deploy an agent. The designer's intention with deploying an agent is for it to assist in building the ideal city. Let $S_0$ denote the state of the environment at the time of its potential introduction. See Figure \ref{fig:minecraft-evolution} for a visual summary of how the contemporary state comes about. %
If the AI is not deployed, then the state of the environment will tend to continue to improve according to $r^*$, as a consequence of continued human activity.

As discussed in Section \ref{sec:no-agent}, our MDP model includes a no-agent transition matrix $P_\emptyset$. In Minecraft, that corresponds to no AI agent being deployed.

\subsection{Why Simple Distributions are Intuitively Catastrophic in Minecraft}\label{subsubsec:two-examples}
In this subsection, we present an informal argument for why if a trajectory is sampled from a simple distribution, and independently of $r^*$, then it should be catastrophic in an intuitive sense in Minecraft. To make the discussion more concrete, we first discuss a prototypically simple distribution --- the uniform distribution --- and why that is intuitively catastrophic. We then explain why all distributions simpler than the uniform distribution ought to be catastrophic in Minecraft. %

\subsubsection{The Uniform Distribution is Simple and Intuitively Catastrophic in Minecraft}

Suppose that at each timestep $S^\perp_t$ is sampled uniformly from $\states$ in Minecraft. Assuming the state space size $|\states|$ has a short description, the uniform distribution is very simple to describe in Python. In  \Cref{fig:simple-distributions}(a), we give an example program when the state space has size $440^{10^{9}}$. This is the case when the grid size is $1000\times1000\times 1000$ and there are 440 block types. Since most states in the game of Minecraft do not contain any desirable structures such as sophisticated buildings or beautiful parks, a uniform sample from Minecraft states, independent of $r^*$, is very unlikely to contain such structures. This is in stark contrast with the baseline trajectory which likely contains many structures the human desires. Indeed, past and future human activity ought to give rise to a highly desirable and sophisticated city. Therefore, relative to $\Vbaseline$, the value $V^\perp$ is intuitively catastrophic. While the value of $V^\perp$ is intuitively catastrophic it is still far better than worst-case value: in Minecraft, $S^\perp_t$ is very unlikely to contain detestable structures such as spider buildings.

To give the reader a richer understanding of what a simple distribution might look like, we include another example in \Cref{fig:simple-distributions} (b). In contrast to the uniform distribution discussed above, this other example places all of its probability mass on a single simple state. This emphasizes that a simple distribution need not be diffuse like the uniform distribution, but can be extremely concentrated.

\begin{figure}[t]
\centering
\begin{minipage}[t]{0.48\textwidth}
\centering

\begin{lstlisting}[style=pynarrow]
def p(h, s):
    return 1 / 440**10**9
\end{lstlisting}
\subcaption{The uniform distribution. This example assumes that $|\states| = 440^{10^9}$.}%
\label{lst:uniform}
\end{minipage}\hfill
\begin{minipage}[t]{0.48\textwidth}
\centering

\begin{lstlisting}[style=pynarrow]
def p(h, s):
    S = (0,) * 10 ** 9 # block 0 repeated
    return int(s == S)
    
\end{lstlisting}
\subcaption{The simple singleton distribution. It places all probability mass on a state with the block $0$ (air) repeated throughout the whole grid. It assumes that the grid size is $10^9$.}
\label{lst:singleton}
\end{minipage}
\caption{Two prototypically simple distributions over Minecraft trajectories.  Each is
implemented as a Python program that takes a partial state trajectory $h \in \states^*$ and state $s \in \states$ and returns a
number in $[0,1]$. Both programs are short, yet they sit at opposite ends
of the diffuse--concentrated spectrum.}
\label{fig:simple-distributions}
\end{figure}

\subsubsection{Why All Simple Distributions Are Intuitively Catastrophic in Minecraft}
In the preceding subsection, we explained why the uniform distribution is intuitively catastrophic in Minecraft. But perhaps there exist other simple distributions that are \textit{not} catastrophic. In this section, we argue informally that this is not the case: all distributions as simple as or simpler than the uniform distribution are intuitively catastrophic in Minecraft. That will imply that orthogonal value is intuitively catastrophic.

Let $L_{\mathrm{uniform}}$ be the length of the program in Figure \ref{fig:simple-distributions}(a), which implements the uniform distribution for the case when $|\states|=440^{10^9}$. We assume that all valid programs start with the code snippet \texttt{def p(h,s):} and end with the code snippet \texttt{return} followed by a value, and so we ignore these code snippets when computing the length of a program. In this case, we have 
$$
L_{\mathrm{uniform}} = 12.
$$

Now the key question is: Is it possible to construct another distribution with $12$ symbols or less that is not catastrophic? It seems very likely that the answer is no. If a distribution is to put high probability on non-catastrophic states, it seems very likely that the program actually needs to describe features that the human desires, and that this would require more than $12$ symbols. 

To understand this idea better, it is helpful to consider the special case of singleton distributions. Suppose one wanted to construct a distribution that places all its probability mass on a noncatastrophic state. Then, that state would need to be described in the Python program, and presumably describing such a state (that may involve beautiful parks, or a sophisticated building) would require more than $12$ symbols.

While simple distributions are intuitively catastrophic, they offer far more than worst-case value: it seems very unlikely a program with $12$ symbols or less can implement a distribution placing significant probability mass on detestable states (those that include spider-buildings). For if a program did place much probability mass on such states, the program would presumably need to include some description of those spider-buildings that would likely require more than $12$ symbols.

All in all, it seems highly likely that all distributions with $12$ symbols or less are intuitively catastrophic in Minecraft, while far better than worst-case. It follows that the orthogonal value, which is the value of the simplest distribution is also intuitively catastrophic, while far better than worst-case value.

\subsubsection{Beyond Python}
What constitutes the simplest distribution depends on the programming language. While our discussion has focused on Python, one may wonder how far the ideas extend to other languages. We briefly discuss this now.

We believe that our argument extends naturally to all commonly used programming languages: if a distribution is sufficiently simple then it will be intuitively catastrophic in Minecraft. Thus, we expect the choice of programming language not to impact our interpretation of orthogonal value. 

More broadly, and as explained in Appendix \ref{sec:simple-distribution-cannot-concentrate-on-rate-states}, there is a formal sense in which the simpler a distribution is, the less it can place mass on complex elements.  This holds for all languages. %
But these formal results are directional and do not rule out languages in which a simple distribution concentrates on specific complex states. Furthermore, what constitutes a complex state is language-dependent. In principle, one can construct programming languages in which the best possible state in Minecraft is the simplest state. In that case, with a technical assumption, there is a simple singleton that places all its mass on the best possible state.  But we consider such cases pathological.

This raises an interesting question for future research: is there a fundamental reason why simple distributions in commonly used languages ought to be catastrophic? Relatedly, is there a fundamental reason why very simple states (in commonly used languages) are unlikely to be desirable?

\section{Related Work}
\label{sec:related-work}

In developing our definition of catastrophic performance, we considered various reference values used in the previous literature. In this section, we situate our work by discussing these reference values and why we did not use them. We hope this will make it easier for the reader to assess the strengths and weaknesses of our definition of catastrophic performance. %

\subsection{Best-Case Value}
Best-case value is a reference value that is ubiquitous in the reinforcement learning literature. Indeed, most notions of regret are defined with respect to best-case value or variants thereof. Here, we just mention a few examples that touch on AI misalignment.

\citet{everittReinforcementLearningCorrupted2017} characterize performance relative to best-case in their study of ``corrupt reward MDPs''. Similarly, in their introduction of a distance function on reward functions, \citet{gleave2021quantifying} derive a bound on regret relative to optimal. \citet{pmlr-v267-fluri25a} use the normalized regret -- namely, the difference from best-case value, divided by the difference between best-case and worst-case value -- as the performance metric in their theory on the perils of optimizing learned reward functions.  \cite{brownValueAlignmentVerification2021a} say a policy is $\epsilon$-value aligned if its value function is uniformly within $\epsilon$ of optimal and study the problem of verifying $\epsilon$-value alignment. %

As discussed in \Cref{sec:best-case-and-worst-case-performance}, characterizing performance relative to best-case does not distinguish between performance that is beneficial and detrimental, let alone catastrophic. 

\subsection{Worst-Case Value}

Sometimes, performance is also characterized using worst-case value, or a variant thereof.  For instance, as mentioned above, \citet{pmlr-v267-fluri25a} characterizes performance relative to the performance of a best-case and a worst-case policy. 

Another variant of worst-case value is worst-case return possible based on the interface, even if not achievable by any policy. This is often used as the zero-point when reporting performance scores in reinforcement learning. See, for example, \citet{peng2018deepmimic, 10.1145/3272127.3275014,abdulhai2025lmrl}.

Finally, we mention \citet{cross2026fragilityvalueimperfectalignment} which define what they refer to as partial catastrophe. In particular, they say that a performance level $V$ is $\nu$-catastrophic if $V \le \nu$. Thus, $\nu$-catastrophic means ``performance is worse or equal to $\nu$''. Thus, their notion is not defined relative to any reference values at all. However, they do define ``catastrophe'' as a performance level equal to worst-case performance.

As discussed in \Cref{sec:best-case-and-worst-case-performance}, worst-case value does not capture the nature and magnitude of the central concern in the alignment literature. 

\subsection{Inaction Value}
Some works define a baseline state or trajectory corresponding to an agent executing a ``no-op'' action or an agent that performs ``no optimization''. Such reference values are akin to our definition of baseline value.

\citet{elmhamdi2024goodhartslawapplicationvalue} and \citet{zhuang2020consequences} develop models of how overoptimization can lead to detrimental outcomes. In their plots, both characterize performance relative to the performance achieved when there is no optimization at all.%

The side-effects literature (see e.g.,  \citet{armstrong2017low}, \citet{krakovna2018measuring}, and \citet{turner2020conservative,turner2020avoiding}) develops methods that incentivize an agent to avoid causing unnecessary changes to the environment. These approaches typically define a baseline state or trajectory, often corresponding to a scenario in which the agent takes no action, and penalize deviations from that baseline. In the side-effects literature, the baseline serves as a reference point for measuring impact and constraining behavior, rather than a reference value against which to evaluate performance. %

In a blog post, \citet{kosoy2021hippocratic} proposes a closely related reference value. The setup considered is the so-called delegative RL setup in which at every timestep the agent can choose to delegate to the human. The baseline proposed is the expected performance of an agent that always delegates to the human.

\subsection{Random Action Value}
A relatively common reference value is the value of an agent that samples actions uniformly at random at each timestep. For instance, in \citet{mnih2015human}, they normalize performance such that $0$ is the performance of an agent that sampled actions uniformly at random, and $1$ is the performance of a human.

In environments where random actions tend not to impact the environment much, this baseline represents performance similar to, but slightly worse than, baseline value. For instance, consider deploying an agent in Minecraft that randomizes uniformly across actions. Such an agent would almost certainly not impact the environment much at all. Expected performance would be slightly below baseline value, but much better than orthogonal value.

It is worth noting that there can, in principle, be environments in which random actions are likely to radically change the state of the environment and generate a very detrimental outcome.  In that case, the random-action baseline may represent performance that is more akin to our definition of orthogonal value.

Because the magnitude of the random action value can vary across environments in this way, we think it is an inappropriate alternative for either baseline or orthogonal value.

\subsection{Unrelated Goal Value}
In writing this paper, before arriving at our definition of orthogonal value, we first worked with another definition. Loosely speaking, it was ``the expected return of a maximally competent agent optimizing a reward function unrelated to human preferences''. We leave out how we formalized ``maximally competent'' and ``unrelated''. We were motivated by the intuition that a maximally competent agent that optimizes an ``arbitrary'' reward function ought to be catastrophic in many contexts.  We used such a definition in an early version of a different paper of ours \citep{marklund2026consequentialist}. 

That definition is problematic because the validity of the definition relies on nonobvious facts about the environment. In particular, it relies on what an agent optimizing an unrelated reward function is likely to achieve. We think it is better to use a definition whose validity does not rely on such properties.

\subsection{Other Reference values}
\citet{laidlaw2024correlated} provide a definition of reward hacking using a baseline policy: a
policy hacks $\hat r$ relative to a chosen reference policy $\pi_{\mathrm{ref}}$ if it
$\hat r$-dominates $\pi_{\mathrm{ref}}$ but is $r^*$-dominated by it. Here, poor performance corresponds to performing worse than the reference policy.

In their framework, $\pi_{\mathrm{ref}}$ is a free parameter. Thus,   performing worse than $\pi_{\mathrm{ref}}$ does not necessarily mean detrimental performance since $\pi_{\mathrm{ref}}$ may be a relatively good policy. Also, their definition of reward hacking is relative to the single baseline $\pi_{\mathrm{ref}}$. Using a single baseline cannot provide a scale: their theory does not distinguish between a policy that performs just barely worse than the reference policy and one that performs significantly worse.

\subsection{Other Relevant Work}

\textbf{Reward hacking:}  The reward hacking literature studies what happens when an agent optimizes a proxy reward $\hat r$ that differs from the true reward $r^*$; see, for instance, \cite{karwowski2023goodhart, manheim2019categorizingvariantsgoodhartslaw}. At a high level, reward hacking happens when the agent performs well according to $\hat r$ and poorly according to $r^*$. Some papers in the reward hacking literature have already been discussed above. 

\citet{skalse2022defining} define a reward function $r^*$ and its proxy $\hat r$ to be \emph{hackable} if there is a pair of policies for which the proxy strictly prefers a strictly $r^*$-inferior alternative. %
But the fact that $r^*$ and $\hat r$ are hackable does not tell us whether an induced policy $\hat \pi$ produces beneficial or detrimental, let alone catastrophic performance.

\textbf{Goal-misgeneralization:} This body of work studies how deployed agents may pursue goals at odds with the goals encoded by the training objective.
\citet{di2022goal} provide a semi-formal definition of goal misgeneralization. They say a policy goal-misgeneralizes if its trajectories are better explained as those of a policy with ``goal-directedness'' instead of a policy without goal-directedness, and it receives low reward according to $r^*$. They formally define goal-directedness but do not formally define what counts as low reward. \citet{shah2022goal} also study goal misgeneralization but do not formally specify any reference value in their study (although in their Monster Gridworld and Cultural Transmission examples, they do show the performance of a random policy for comparison). \citet{abdelsadek2025mitigating} compare expected reward maximization with a regret minimization method for mitigating goal-misgeneralization. They use regret (relative to an optimal policy) as their performance metric.

\section{Conclusion}
Previous theoretical works often characterize agent performance relative to the best-case and, sometimes, worst-case value. This paper introduced two alternative reference values, baseline and orthogonal value, in order to define a meaningful notion of catastrophic performance. Unlike best-case value, baseline value provides a way to distinguish between detrimental performance and mere suboptimality. Compared to worst-case value, orthogonal value more accurately reflects the nature and magnitude of the central concern of the alignment community. We believe these new reference values will prove more useful in analyzing catastrophic risk from misaligned AI.

\section*{Acknowledgments}
This work was supported by grants from the Huxley Foundation and the UK AI Security Institute. We thank Jonathan Carr, Ryan Carey, Alex Cloud, Michael Cohen, John Duchi, Blake Elias, Benlin Gan, Stephane Hatgis-Kessell, Ramesh Johari, Cameron Jordan, Saurabh Kumar, Smitha Milli, Stuart Russell, Karim Abdel Sadek, Stewy Slocum, Eric Tang, Bryce Woodworth, Renyuan Xu, Wanqiao Xu, and Yifan Zhu for helpful comments and discussions. We also thank students of the MS\&E 338/CS 338 course for valuable discussions. Finally, we used AI tools to check and improve the writing of the paper and find relevant previous work. It also assisted in the writing of proofs in the appendix, and provided key ideas for the proof of \Cref{prop:simple-dist-cannot-concentrate-on-rate-elements-conditional} in the appendix.

\bibliographystyle{plainnat}  %
\bibliography{references}

@book{russell2019human,
  title={Human compatible: {AI} and the problem of control},
  author={Russell, Stuart},
  year={2019},
  publisher={Penguin UK}
}

@incollection{omohundro2018basic,
  title={The basic {AI} drives},
  author={Omohundro, Stephen M},
  booktitle={Artificial intelligence safety and security},
  pages={47--55},
  year={2018},
  publisher={Chapman and Hall/CRC}
}

@article{skalse2022defining,
  title={Defining and characterizing reward gaming},
  author={Skalse, Joar and Howe, Nikolaus and Krasheninnikov, Dmitrii and Krueger, David},
  journal={Advances in Neural Information Processing Systems},
  volume={35},
  pages={9460--9471},
  year={2022}
}

@article{laidlaw2024correlated,
  title={Correlated proxies: A new definition and improved mitigation for reward hacking},
  author={Laidlaw, Cassidy and Singhal, Shivam and Dragan, Anca},
  journal={arXiv preprint arXiv:2403.03185},
  year={2024}
}

@misc{manheim2019categorizingvariantsgoodhartslaw,
  title={Categorizing Variants of {G}oodhart's Law},
  author={David Manheim and Scott Garrabrant},
  year={2019},
  eprint={1803.04585},
  archivePrefix={arXiv},
  primaryClass={cs.AI},
  url={https://arxiv.org/abs/1803.04585}
}

@misc{elmhamdi2024goodhartslawapplicationvalue,
  title={On {G}oodhart's law, with an application to value alignment},
  author={El-Mahdi El-Mhamdi and Lê-Nguyên Hoang},
  year={2024},
  eprint={2410.09638},
  archivePrefix={arXiv},
  primaryClass={stat.ML},
  url={https://arxiv.org/abs/2410.09638}
}

@article{peng2018deepmimic,
  title={Deepmimic: Example-guided deep reinforcement learning of physics-based character skills},
  author={Peng, Xue Bin and Abbeel, Pieter and Levine, Sergey and Van de Panne, Michiel},
  journal={ACM Transactions On Graphics (TOG)},
  volume={37},
  number={4},
  pages={1--14},
  year={2018},
  publisher={ACM New York, NY, USA}
}

@inproceedings{di2022goal,
  title={Goal misgeneralization in deep reinforcement learning},
  author={Di Langosco, Lauro Langosco and Koch, Jack and Sharkey, Lee D and Pfau, Jacob and Krueger, David},
  booktitle={International Conference on Machine Learning},
  pages={12004--12019},
  year={2022},
  organization={PMLR}
}

@article{shah2022goal,
  title={Goal misgeneralization: Why correct specifications aren't enough for correct goals},
  author={Shah, Rohin and Varma, Vikrant and Kumar, Ramana and Phuong, Mary and Krakovna, Victoria and Uesato, Jonathan and Kenton, Zac},
  journal={arXiv preprint arXiv:2210.01790},
  year={2022}
}

@inproceedings{abdelsadek2025mitigating,
  title={Mitigating Goal Misgeneralization via Minimax Regret},
  author={Abdel Sadek, Karim and Farrugia-Roberts, Matthew and Anwar, Usman and Erlebach, Hannah and Schroeder de Witt, Christian and Krueger, David and Dennis, Michael},
  booktitle={Reinforcement Learning Conference (RLC)},
  year={2025},
  note={arXiv:2507.03068}
}

@article{hadfield2016cooperative,
  title={Cooperative inverse reinforcement learning},
  author={Hadfield-Menell, Dylan and Russell, Stuart J and Abbeel, Pieter and Dragan, Anca},
  journal={Advances in neural information processing systems},
  volume={29},
  year={2016}
}

@article{zhuang2020consequences,
  title={Consequences of misaligned {AI}},
  author={Zhuang, Simon and Hadfield-Menell, Dylan},
  journal={Advances in Neural Information Processing Systems},
  volume={33},
  pages={15763--15773},
  year={2020}
}

@article{karwowski2023goodhart,
  title={Goodhart's law in reinforcement learning},
  author={Karwowski, Jacek and Hayman, Oliver and Bai, Xingjian and Kiendlhofer, Klaus and Griffin, Charlie and Skalse, Joar},
  journal={arXiv preprint arXiv:2310.09144},
  year={2023}
}

@article{krakovna2018measuring,
  title={Measuring and avoiding side effects using relative reachability},
  author={Krakovna, Victoria and Orseau, Laurent and Kumar, Ramana and Martic, Miljan and Legg, Shane},
  journal={arXiv preprint arXiv:1806.01186},
  year={2018}
}

@inproceedings{turner2020conservative,
  title={Conservative Agency via Attainable Utility Preservation},
  author={Turner, Alexander Matt and Hadfield-Menell, Dylan and Tadepalli, Prasad},
  booktitle={Proceedings of the AAAI/ACM Conference on AI, Ethics, and Society},
  pages={385--391},
  year={2020}
}

@inproceedings{turner2020avoiding,
  title={Avoiding Side Effects By Considering Future Tasks},
  author={Turner, Alexander Matt and Ratzlaff, Neale and Tadepalli, Prasad},
  booktitle={Advances in Neural Information Processing Systems},
  year={2020}
}

@article{armstrong2017low,
  title={Low Impact Artificial Intelligences},
  author={Armstrong, Stuart and Levinstein, Benjamin},
  journal={arXiv preprint arXiv:1705.10720},
  year={2017}
}

@InProceedings{pmlr-v267-fluri25a,
  title = 	 {The Perils of Optimizing Learned Reward Functions: Low Training Error Does Not Guarantee Low Regret},
  author =       {Fluri, Lukas and Lang, Leon and Abate, Alessandro and Forr\'{e}, Patrick and Krueger, David and Skalse, Joar Max Viktor},
  booktitle = 	 {Proceedings of the 42nd International Conference on Machine Learning},
  pages = 	 {17306--17377},
  year = 	 {2025},
  editor = 	 {Singh, Aarti and Fazel, Maryam and Hsu, Daniel and Lacoste-Julien, Simon and Berkenkamp, Felix and Maharaj, Tegan and Wagstaff, Kiri and Zhu, Jerry},
  volume = 	 {267},
  series = 	 {Proceedings of Machine Learning Research},
  month = 	 {13--19 Jul},
  publisher =    {PMLR},
  url = 	 {https://proceedings.mlr.press/v267/fluri25a.html}
}

@article{shah2019preferences,
  title={Preferences implicit in the state of the world},
  author={Shah, Rohin and Krasheninnikov, Dmitrii and Alexander, Jordan and Abbeel, Pieter and Dragan, Anca},
  journal={arXiv preprint arXiv:1902.04198},
  year={2019}
}

@article{garrabrant2016logical,
  title={Logical induction},
  author={Garrabrant, Scott and Benson-Tilsen, Tsvi and Critch, Andrew and Soares, Nate and Taylor, Jessica},
  journal={arXiv preprint arXiv:1609.03543},
  year={2016}
}

@article{yudkowsky2008artificial,
  title={Artificial intelligence as a positive and negative factor in global risk},
  author={Yudkowsky, Eliezer},
  journal={Global catastrophic risks},
  volume={1},
  number={303},
  pages={184},
  year={2008},
  publisher={New York}
}

@misc{cross2026fragilityvalueimperfectalignment,
      title={Fragility of Value under Imperfect Alignment}, 
      author={Winter Cross and Léo Cymbalista and Alfred Harwood and Jose Faustino},
      year={2026},
      eprint={2607.28881},
      archivePrefix={arXiv},
      primaryClass={cs.AI},
      url={https://arxiv.org/abs/2607.28881}, 
}

@inproceedings{brownValueAlignmentVerification2021a,
  title = {Value {{Alignment Verification}}},
  booktitle = {Proceedings of the 38th {{International Conference}} on {{Machine Learning}}},
  author = {Brown, Daniel S. and Schneider, Jordan and Dragan, Anca and Niekum, Scott},
  year = 2021,
  month = jul,
  pages = {1105--1115},
  publisher = {PMLR},
  issn = {2640-3498},
  url = {https://proceedings.mlr.press/v139/brown21a.html},
  urldate = {2026-09-11},
  langid = {english}
}

@inproceedings{
gleave2021quantifying,
title={Quantifying Differences in Reward Functions},
author={Adam Gleave and Michael D Dennis and Shane Legg and Stuart Russell and Jan Leike},
booktitle={International Conference on Learning Representations},
year={2021},
url={https://openreview.net/forum?id=LwEQnp6CYev}
}

@article{mnih2015human,
  title={Human-level control through deep reinforcement learning},
  author={Mnih, Volodymyr and Kavukcuoglu, Koray and Silver, David and Rusu, Andrei A and Veness, Joel and Bellemare, Marc G and Graves, Alex and Riedmiller, Martin and Fidjeland, Andreas K and Ostrovski, Georg and others},
  journal={nature},
  volume={518},
  number={7540},
  pages={529--533},
  year={2015},
  publisher={Nature Publishing Group UK London}
}

@inproceedings{everittReinforcementLearningCorrupted2017,
  title = {Reinforcement {{Learning}} with a {{Corrupted Reward Channel}}},
  booktitle = {Proceedings of the {{Twenty-Sixth International Joint Conference}} on {{Artificial Intelligence}}},
  author = {Everitt, Tom and Krakovna, Victoria and Orseau, Laurent and Legg, Shane},
  year = 2017,
  month = aug,
  pages = {4705--4713},
  publisher = {International Joint Conferences on Artificial Intelligence Organization},
  address = {Melbourne, Australia},
  doi = {10.24963/ijcai.2017/656},
  url = {https://www.ijcai.org/proceedings/2017/656},
  urldate = {2026-09-10},
  isbn = {978-0-9992411-0-3},
  langid = {english}
}

@article{10.1145/3272127.3275014,
author = {Peng, Xue Bin and Kanazawa, Angjoo and Malik, Jitendra and Abbeel, Pieter and Levine, Sergey},
title = {SFV: reinforcement learning of physical skills from videos},
year = {2018},
issue_date = {December 2018},
publisher = {Association for Computing Machinery},
address = {New York, NY, USA},
volume = {37},
number = {6},
issn = {0730-0301},
url = {https://doi.org/10.1145/3272127.3275014},
doi = {10.1145/3272127.3275014},
journal = {ACM Trans. Graph.},
month = dec,
articleno = {178},
numpages = {14}
}

@inproceedings{
abdulhai2025lmrl,
title={{LMRL} Gym: Benchmarks for Multi-Turn Reinforcement Learning with Language Models},
author={Marwa Abdulhai and Isadora White and Charlie Victor Snell and Charles Sun and Joey Hong and Yuexiang Zhai and Kelvin Xu and Sergey Levine},
booktitle={Forty-second International Conference on Machine Learning},
year={2025},
url={https://openreview.net/forum?id=hmGhP5DO2W}
}

@article{marklund2026consequentialist,
  title={Consequentialist Objectives and Catastrophe},
  author={Marklund, Henrik and Infanger, Alex and Van Roy, Benjamin},
  journal={arXiv preprint arXiv:2603.15017},
  year={2026}
}

@misc{kosoy2021hippocratic,
  author       = {Kosoy, Vanessa},
  title        = {The Hippocratic Principle},
  year         = {2021},
  month        = mar,
  howpublished = {AI Alignment Forum},
  url          = {https://www.alignmentforum.org/posts/dPmmuaz9szk26BkmD/vanessa-kosoy-s-shortform?commentId=jMitxvhFceaheD5zb},
  note         = {Last edited March 17, 2021},
}

@book{liIntroductionKolmogorovComplexity2008,
  title = {An {{Introduction}} to {{Kolmogorov Complexity}} and {{Its Applications}}},
  author = {Li, Ming and Vit{\'a}nyi, Paul},
  year = 2008,
  series = {Texts in {{Computer Science}}},
  publisher = {Springer},
  address = {New York, NY},
  doi = {10.1007/978-0-387-49820-1},
  url = {https://link.springer.com/10.1007/978-0-387-49820-1},
  urldate = {2026-09-13},
  copyright = {https://www.springernature.com/gp/researchers/text-and-data-mining},
  isbn = {978-0-387-33998-6 978-0-387-49820-1},
  langid = {english}
}

@book{shenComputableFunctions2002a,
  title = {Computable {{Functions}}},
  author = {Shen, A. and Vereshchagin, N.},
  translator = {Dubrovskii, V.},
  year = 2002,
  month = dec,
  series = {The {{Student Mathematical Library}}},
  volume = {19},
  publisher = {American Mathematical Society},
  address = {Providence, Rhode Island},
  doi = {10.1090/stml/019},
  url = {https://www.ams.org/stml/019},
  urldate = {2026-09-13},
  isbn = {978-1-4704-2133-5 978-0-8218-2732-1},
  langid = {english}
}

@book{boolosComputabilityLogic2007,
  title = {Computability and {{Logic}}},
  author = {Boolos, George S. and Burgess, John P. and Jeffrey, Richard C.},
  year = 2007,
  edition = {5},
  publisher = {Cambridge University Press},
  address = {Cambridge},
  doi = {10.1017/CBO9780511804076},
  url = {https://www.cambridge.org/core/books/computability-and-logic/440B4178B7CBF1C241694233716AB271},
  urldate = {2026-10-07},
  isbn = {978-0-521-87752-7}
}

\appendix

\section{Algorithmic Information Theory Background}
\label{app:ait-background}

The goal of this section is to formally define the computability and universality of the interpreter $U:\Sigma^*\times\mathcal X\rightharpoonup\mathcal X$ described in \Cref{subsec:complexity-preliminaries}. To this end, we will first fix a model of computation based on a multi-tape Turing machine.

\subsection{Multitape Turing Machine}
We will take as our model of computation a Turing machine with four tapes:
\begin{enumerate}
    \item read-only tape I,
    \item read-only tape II,
    \item memory tape,
    \item append-only output tape.
\end{enumerate}
Recall that $\Sigma$ is a finite set whose elements we refer to as program symbols. All tapes have $\Sigma$ as their alphabet.

We choose this model to match a standard Python or Java interpreter of the following sort. The Python interpreter takes an input program (read-only tape I), and may take an additional input string (read-only tape II). As the interpreter runs the program, it uses memory to facilitate computation (memory tape) and writes output to the user (append-only output tape).

We will refer to a Turing machine $M$ as a single-input tape Turing machine if its operation is independent of the second input tape.

\subsection{Computable Functions and the Interpreter}

Given our multi-tape Turing machine, we are almost in position to define computability and universality of the interpreter $U$ from \Cref{subsec:complexity-preliminaries}. Let us begin with computability. A function is typically called computable if some Turing machine computes it. The interpreter $U$, however, has domain $\Sigma^*\times \mathcal X$ and codomain $\mathcal X$, whereas our Turing machine only reads and write elements in $\Sigma^*$. 

To bridge this gap, we will make use of a bijective decoder $d_{\mathcal X}:\Sigma^*\to\mathcal X$. Given this decoder, our path to defining the computability and universality of $U$ is as follows. We will first define computability for partial functions $f:\Sigma^*\times \Sigma^*\rightharpoonup\Sigma^*$. Then, we will say the interpreter $U$ is computable if the induced partial function $\tilde U:\Sigma^*\times\Sigma^*\rightharpoonup\Sigma^*$ with  $\tilde{U}(p,x) = d_{\mathcal{X}}^{-1}\bigl(U(p, d_{\mathcal{X}}(x))\bigr)$ (with $\tilde{U}(p,x)$ undefined whenever $U(p, d_{\mathcal{X}}(x))$ is undefined), is computable. 

We will similarly first define universality on computable partial functions from $\Sigma^*\times\Sigma^*\rightharpoonup\Sigma^*$ and then extend it to functions from $\Sigma^*\times \mathcal X\rightharpoonup \mathcal X$ using the decoder $d_{\mathcal X}$.

We now formally develop the theory outlined above. 

\begin{definition}
    The partial function $f: \Sigma^* \times \Sigma^* \rightharpoonup \Sigma^*$ is said to be implemented by the Turing machine $M$ if for all $x,x' \in \Sigma^*$ 
    \begin{enumerate}
        \item $M$ halts on $(x,x')$ if and only if $f$ is defined on $(x,x')$,
        \item and if, $M$ halts on $(x,x')$, the output is $f(x,x')$.
    \end{enumerate}
    Similarly, a partial function $h: \Sigma^* \rightharpoonup \Sigma^*$ is implemented by a single-input tape Turing machine if analogous conditions hold.
\end{definition}

\begin{definition}
    A partial function $f: \Sigma^* \times \Sigma^* \rightharpoonup \Sigma^*$ is computable if there exists a Turing machine that implements it. Similarly, a partial function $h: \Sigma^* \rightharpoonup \Sigma^*$ is computable if a single-input tape Turing machine implements it.
\end{definition}

\begin{definition} Let $f:\Sigma^*\times \Sigma^*\rightharpoonup \Sigma^*$ be a computable partial function. If for all computable partial functions $h:\Sigma^*\rightharpoonup\Sigma^*$, there exists a $p\in\Sigma^*$ such that
$$
f(p,\cdot)=h(\cdot),
$$
then $f$ is called \textit{universal}. 
\end{definition}

\begin{remark} Universality is sometimes used to describe Turing machines as opposed to computable functions (see, e.g., \citet{liIntroductionKolmogorovComplexity2008}). Our approach of using it as a modifier on computable functions is also seen in the literature (see, e.g., \citet{shenComputableFunctions2002a} and \citet{boolosComputabilityLogic2007}). 

\end{remark}

We will now extend our definitions of computability and universality to functions which have domain $\Sigma^*\times \mathcal X$ and codomain $\mathcal X$. Recall that our decoder $d_{\mathcal X}:\Sigma^*\to\mathcal X$ is a bijective map. 

\begin{definition} The partial function $g:\Sigma^*\times \mathcal X\rightharpoonup\mathcal X$ is computable if the partial function $f:\Sigma^*\times \Sigma^*\rightharpoonup \Sigma^*$ defined by 
$$
f(x,y) = \begin{cases} d^{-1}_{\mathcal X}(g(x,d_{\mathcal X}(y))) & (x,d_{\mathcal X}(y))\in \text{dom}(g)\\
\text{undefined} & \text{otherwise}.
\end{cases}
$$
is computable. Further, $g$ is called universal if $f$ is universal. 
\end{definition}

\begin{definition}
    An interpreter is a universal computable partial function $U:\Sigma^*\times \mathcal X\rightharpoonup\mathcal X$. 
\end{definition}

\section{The Prior:  The Agent's or the Designer's  Uncertainty?}
\label{app:why-designer-uncertinty-rather-than-agent-uncertainty}
It is quite common in previous work to treat the prior as representing the agent's beliefs. In particular, it is common to take the prior to be a probability distribution that the designer equips the agent with. 

In this case, the prior is simply an algorithmic component chosen by the designer. And since the prior is part of the algorithm that induces the agent policy, it no longer makes much sense to evaluate the agent policy \textit{relative} to the prior. Instead, in this case, the prior should be regarded as part of the object being evaluated.

So what should the prior represent? Ultimately, we want to enable theoretical analysis of the conditions that give rise to misaligned AI agents producing catastrophic outcomes. For that purpose, we want to evaluate an agent policy relative to the beliefs of the designer (or the beliefs of some third party) and not the beliefs of the agent. That is why we take the prior to encode uncertainty stemming from the designer.

\section{The POMDP Formulation}
\label{app:pomdp-formulation}
Our definitions of baseline value, orthogonal value, and $\eta$-catastrophic performance extend naturally to more general settings. In this section, we illustrate this by extending the framework to partially observable environments using the \textit{partially observed} MDP (POMDP) formalism.

\subsection{Preliminaries}
A partially observed MDP is a tuple $(\states,\actions,\observations,P,F,s_0, \gamma, r)$, consisting of an MDP $(\states, \actions, P, s_0, \gamma, r)$ equipped with an observation space $\observations$ and \textit{observation kernel} $F : \states \to \Delta_{\observations}$.

As before, we extend the transition kernel to include an additional transition matrix $P_\emptyset$.  The subscript $\emptyset$ indicates that no agent is deployed and we use this matrix to model dynamics when the agent is not deployed.  Absent the agent's influence, the state transitions according to $P_\emptyset$.  On the other hand, with an active agent that selects an action $a \in \actions$, the state transitions according to $P_a$.

The \textit{true POMDP} is a random tuple $(\states, \actions, \observations, P^*, F^*, S_0, r^*, \gamma)$.  Here, $(\states, \actions, \observations, \gamma)$ is deterministic, but $(P^*, F^*, S_0, r^*)$ is random. %

\subsection{Baseline and Orthogonal Value}
The definitions are exactly as in the main text. For extra clarity, we write them out. Let $H^\emptyset$ be the state trajectory generated according to the transition matrix $P_\emptyset^*$.  As before, we refer to $H^\emptyset$ as the \textit{baseline trajectory}.
\begin{definition}
The baseline value is 
$$
\Vbaseline = \E \left[r^*_\gamma(H^\emptyset) \right],
$$
where $H^\emptyset$ is the baseline trajectory.
\end{definition}

\begin{definition}
The orthogonal value is 
$$
\vorthogonal = \E \left[r^*_\gamma(H^\perp) \right],
$$
where $H^\perp$ is a random trajectory of states that is independent of $r^*$ and has the simplest distribution. %
\end{definition}

\section{Proofs for \Cref{sec:example-gridworld} (Gridworld Example)}
\label{app:gridworld-proofs}
For these proofs, it will be useful to extend the definition of $M$ and $S_0$ to allow for the case $p=0$ and $p=1$. 
\begin{lemma}
If $p = 1$, then
$$
    \E[r^*(S_0)] = 0.
$$
\end{lemma}
\begin{proof}
    Since $p=1$, we have that $M=n$ with probability 1 and $M$ is thus deterministic. Therefore, $S_0 \perp r^*$. Since $\E[r^*(s)] = 0$ for all $s \in \states$, it follows that $\E[r^*(S_0)] = 0$.
\end{proof}

\begin{lemma}
Let $S_0'$ be a state sampled independently of $S_0$, in the same way as $S_0$ but with parameter $p' \in [0,1]$ instead. If $p < p'$, then
    $$
        \E[r^*(S_0)] > \E[r^*(S_0')].
    $$
\end{lemma}
\begin{proof}
    Let $M'$ be the random distance used for sampling $S_0'$. 
    
    First note that the distribution of $M'$ stochastically dominates $M$. Therefore, $$\E\left[e^{-d^2(S^+, S_0)/2\sigma^2}\right] > \E\left[e^{-d^2(S^+, S_0')/2\sigma^2}\right].$$

    Since $S^-$ is uniform on $\states$ and the boundary wraps around, $e^{-d^2(S^-, s)/2\sigma^2}$ has the same distribution for all states $s \in \states$.
    
    Since $S_0$ and $S_0'$ are both independent of $S^-$, we therefore have that
    $$\E\left[e^{-d^2(S^-, S_0)/2\sigma^2}\right] = \E\left[e^{-d^2(S^-, S_0')/2\sigma^2}\right].$$

    The result follows.
\end{proof}
\begin{proposition}
    If $p\in[0,1)$, we have that 
    $$
    \E [r^*(S_0)]>0
    $$
\end{proposition}
\begin{proof}
    The result follows immediately from the preceding lemmas.
\end{proof}

\begin{proposition}
    Suppose $q\in[0,1)$. Then $$
\E[r^*(S_{t+1}^\emptyset)] > \E[r^*(S_0)].
$$
\end{proposition}
\begin{proof}   
Let $r_-^*$ and $r_+^*$ be defined for all $s$ as follows:
\begin{align*}
r_-^*(s) &= e^{-d^2 (S^-, s)/2\sigma^2} \\
r_+^*(s) &= e^{-d^2 (S^+, s)/2\sigma^2}
\end{align*}

Then, for all $t$ we have that 
$$\E[r^*_-(S_0)] = \E[r^*_-(S^\emptyset_{t+1})].$$

And for all $t$
$$
\E[r^*_+(S_0)] < \E[r^*_+(S^\emptyset_{t+1})].
$$
The result follows since $r^*(s)=r^*_+(s)-r_-^*(s)$.

\end{proof}

\section{A Simple Distribution Cannot Concentrate on Rare Elements}
\label{sec:simple-distribution-cannot-concentrate-on-rate-states}

\subsection{Preliminaries}

We make a weak assumption about the programming language.
\begin{assumption}
    For all computable functions $f: \mathcal{X} \to \mathcal{X}$, there exists $c_f \ge 0$ such that for all $x \in \mathcal{X}$,
    $$
        C(f(x)) \le C(x) + c_f.
    $$
\end{assumption}

Recall that $\mathcal X$ comes equipped with a computable enumeration. This allows us to define a computable ordering on $\mathcal X$ based on which element comes first in the enumeration. 

In what follows, we will make use of the following notion of density of a subset $X$ of $\mathcal X$.

\begin{definition}
The program density of $X \subseteq \mathcal{X}$ at program length $k \in \mathbb{Z}_{\ge 0}$ is
    $$
        \phi_k(X) = \frac{|\{p \in \Sigma^{\le k} : U(p, \epsilon) \in X \} |} {|\Sigma^{\le k}|}.
    $$
\end{definition}

In the main text preliminaries, we defined what a computable function is. We now define what it means for a distribution $p \in \Delta_{\mathcal{X}}$ on $\mathcal{X}$ to be computable. To do so, let us treat $p$ as a function from $\mathcal{X}$ to $[0,1]$. We say that $p$ is computable if the function is computable. %

\begin{restatable}{definition}{ctilde}
 For all computable $p \in \Delta_{\mathcal{X}}$, let
 $$\tilde c(p) = C(p) + 2 \log C(p) + c,$$   
 where $c$ is the constant from \Cref{lemma:index-bound-complexity}.%
\end{restatable}

In our lemma and proofs we will write $\log$ to indicate the logarithm in base $|\Sigma|$.

\subsection{Lemma: Complex States Must Come Late in the Probability Ordering}
\begin{definition} 
(probability index)
For a computable probability distribution $p\in\Delta_{\mathcal X}$, the index function $\mathrm{index}_p:\mathrm{supp}(p)\to \mathbb{Z}_{\ge 1}$ is defined such that for all $x\in\text{supp}(p)$, $\mathrm{index}_p(x)$ returns the index of $x$ in a list of $\text{supp}(p)$ sorted by probabilities under $p$ in descending order, with ties broken by the original ordering that $\mathcal X$ comes equipped with.
\end{definition}

\begin{restatable}{lemma}{indexboundcomplexity}\label{lemma:index-bound-complexity}
There exists $c \ge 0 $ such that for all computable $p \in \Delta_\mathcal{X}$ and all $x \in \mathrm{supp}(p)$,
   $$
        C(x) \le \log \mathrm{index}_p(x) + C(p) + 2 \log C(p) + c.
   $$    
\end{restatable}
\begin{proofsketch} 
   Let $x \in \mathrm{supp}({p})$. We will give a program to specify $x$ with length at most $\log \mathrm{index}_p(x) + C(p) + 2 \log C(p) + c$. The result will follow. 

   The program will include $p$ as a subroutine and $\mathrm{index}_p(x)$ as a hard-coded constant. The program will enumerate the ordered elements of $p$ and pick the $\mathrm{index}_p(x)$th one. The length $L$ of such a program will be upperbounded by 
   $$
         L \le C(p)+\lceil \log\mathrm{index}_p(x) \rceil + 2\log(C(p))+c'.
   $$
   Let us explain each term in turn. The term $C(p)$ shows up because that number of symbols is sufficient to specify the subroutine for $p$. The term $\lceil \log\mathrm{index}_p(x) \rceil$ shows up because that number of symbols is sufficient to hard-code $\mathrm{index}_p(x)$. The term $2\log(C(p))$ is an upperbound bound on the number of symbols necessary to delimit when the subroutine $p$ ends and harded coded $\mathrm{index}_p(x)$ begins. Finally, the constant $c'$ corresponds to overhead in the enumeration of ordered elements that is independent of $p$ and $\mathrm{index}_p(x)$. We let $c = c' + 1$ so we can ignore the ceiling in the complexity of $\mathrm{index}_p(x)$. The result follows.

    Because $p$ may have infinite support, the enumeration and selection procedure above needs to be handled carefully. We spell out the exact algorithm in Appendix \ref{sec:DetailedProofs}.
\end{proofsketch}

\begin{remark}
    The preceding lemma implies that complex elements must have high probability index.
\end{remark}

Next, we state a corollary that we don't actually use but is helpful for building intuition.
\begin{corollary}
There exists $c \ge 0 $ such that for all computable $p \in \Delta_\mathcal{X}$ and all $x \in \mathrm{supp}(p)$,
   $$
        p(x) \le \frac{1}{|\Sigma|^{C(x) - C(p) - 2 \log C(p) - c}},
   $$ 
where $c$ is the constant from the preceding lemma.
\end{corollary}
\begin{proof}
Rearranging terms we have the following:
    \begin{align*}
        \log \mathrm{index}_p (x) &\ge C(x) - C(p) - 2 \log C(p) - c \\
        \implies \mathrm{index}_p (x) &\ge |\Sigma|^{C(x) - C(p) - 2 \log C(p) - c} \\
        \implies p (x) &\le \frac{1}{|\Sigma|^{C(x) - C(p) - 2 \log C(p) - c}}.
    \end{align*}
The last step follows because if an element $x$ has $\mathrm{index}_p(x)$ then its probability must be less than $1/\mathrm{index}_p(x)$.
\end{proof}

\subsection{Main Result}

\begin{proposition}
\label{prop:simple-dist-cannot-concentrate-on-rate-elements-conditional}
Let $X \subseteq \mathcal{X}$ be a random set and $Z$ an $X$-measurable random variable.  There exists $c \ge 0$ such that for all computable distributions $p \in \Delta_\mathcal{X}$,
    $$
        \Pr \left( \widetilde Y \in X  \big| Z\right) \le \sup_{k} \E[\phi_{k}(X) \big| Z] |\Sigma|^{C(p) + 2 \log C(p) + c},
    $$
    with probability 1, where $\widetilde Y$ is sampled from $p$ independently of $X$. The constant $c$ is the constant from \Cref{lemma:index-bound-complexity} plus 1.
\end{proposition}
\begin{proof} Let $n_p=|\text{supp}(p)|\in\mathbb Z_{\geq 0}\cup \{\infty\}$. For all $n\in\{1,2,...,n_p\}$ (interpreted as $\mathbb Z_{\geq 1}$ if $n_p=\infty$), let $P_n=\{x:\index_p(x)\leq n\}$ be the first $n$ elements according to the probability ordering induced by $p$. By \Cref{lemma:index-bound-complexity}, we have that for each element $x \in P_n$,  the complexity is bounded as $C(x) \le \log n+\tilde c(p)$, where $\tilde c(p)= C(p) + 2 \log C(p) + c$. Thus, 
$$
    P_n \subseteq \mathcal X_{\lfloor \log n  + \tilde c(p)\rfloor},
$$
where $\mathcal X_k$ is the set of elements of $\mathcal X$ with complexity $k$ or less.

What fraction of elements in $P_n$ are also in $X$? We compute a bound
\begin{align}
     \frac{| P_n \cap X |}{|P_n|}
     &\overset{(a)}{\le} \frac{| \mathcal X_{\lfloor \log n + \tilde c(p)\rfloor} \cap X   |}{|P_n|}\nonumber\\
     &\overset{(b)}{\le} \frac{\phi_{\lfloor \log n  + \tilde c(p)\rfloor}(X) |\Sigma^{\le \lfloor \log n  + \tilde c(p)\rfloor}|}{|P_n|}\nonumber  \\
               &\overset{(c)}{\le} \phi_{\lfloor \log n  + \tilde c(p)\rfloor}(X)\frac{|\Sigma|^{\log n + \tilde c(p) + 1}}{n}\nonumber  \\
               &= \phi_{\lfloor \log n  + \tilde c(p)\rfloor}(X)|\Sigma|^{\tilde c(p) + 1}
               \label{eq:inequality-for-uniform}
\end{align}
Step (a) follows since $P_n \subseteq \mathcal{X}_{\lfloor \log n  + \tilde c (p)\rfloor}$. Step (b) follows since each element in $\mathcal{X}_{\lfloor \log n  + \tilde c(p)\rfloor}$ as at least one program associated with it in $\Sigma^{\lfloor \log n  + \tilde c(p)\rfloor}$ and distinct elements have distinct programs associated with them. Step (c) follows since $|\Sigma^{\leq k}| = \sum_{j=0}^k |\Sigma|^j \leq |\Sigma|^{k+1}$, and since $|P_n| = n$.

Finally, we will refer to each of $P_1$, $P_2$, and so on, as prefixes. It is a known fact that a distribution is equivalent to a mixture of uniform distributions on these prefixes. That is, $\tilde Y$ can be written as
$$
\tilde Y = U_W
$$
where $W$ is a random variable with support $\{1,2,...,n_p\}$ and $(U_n:n\in\{1,...,n_p\})$ is a collection of uniform random variables with $U_n$ uniform on $P_n$. Here, the $U_n$ and $W$ are mutually independent. Note that the $U_n$ and $W$ are also independent of $X$. It follows that,
\begin{align*}
    \Pr(\tilde Y\in X | Z) &= \sum_{n\in\text{supp}(W)}^{}\Pr(W=n|Z)\Pr(U_n\in X|Z,W=n)\\
    &=\sum_{n\in\text{supp}(W)}^{}\Pr(W=n)\E[\1(U_n\in X)|Z,W=n]\\
    &=\sum_{n\in\text{supp}(W)}^{}\Pr(W=n)\E[\1(U_n\in X)|Z]\\
    &=\sum_{n\in\text{supp}(W)}^{}\Pr(W=n)\E[\E[\1(U_n\in X)|X,Z]|Z]\\
    &\overset{(a)}{\leq} \sum_{n\in\text{supp}(W)}^{}P(W=n)\E[\phi_{\lfloor \log n + \tilde c(p)\rfloor}(X)|\Sigma|^{\tilde c(p)+ 1}|Z]\\
    &\leq \sup_{n\in\text{supp}(W)} \E[\phi_{\lfloor \log n  + \tilde c(p)\rfloor}(X) | Z] |\Sigma|^{\tilde c(p)+1} \\
    &\leq \sup_{k\in\mathbb Z_{\geq 0}} \E[\phi_k(X) | Z] |\Sigma|^{\tilde c(p)+1},
\end{align*}
where (a) follows from \eqref{eq:inequality-for-uniform}.     
\end{proof}

\begin{proposition}
\label{lm:simple-dist-cannot-concentrate-on-rate-elements}
Let $X \subseteq \mathcal{X}$ be a random set. There exists $c \ge 0$ such that for all computable distributions $p \in \Delta_\mathcal{X}$,
    $$
        \Pr \left( \widetilde Y \in X  \right) \le \sup_{k} \E[\phi_{k}(X)] |\Sigma|^{C(p) + 2 \log C(p) + c},
    $$
    where $\widetilde Y$ is sampled from $p$ independently of $X$. The constant $c$ is the constant from \Cref{lemma:index-bound-complexity} plus 1.
\end{proposition}
\begin{proof}
    Follows immediately from our previous result.
\end{proof}

\subsection{Proof Details}\label{sec:DetailedProofs}

\begin{proof}[Proof details for \Cref{lemma:index-bound-complexity}]
The proof sketch left out the details of the enumeration and selection algorithm. We discuss these details now.

We loop through all elements $x_1,x_2,...$ using the computable enumeration of $\mathcal X$ and compute $p(x_1),p(x_2), ...$ (this is possible because we assume $p(z)$ is rational for all $z\in\mathcal X$). At each step $k$, we compute an ordered list $(y^{(k)}_1,..., y^{(k)}_k)$ by sorting $(x_1,..., x_k)$ by their probabilities, breaking ties via the original enumeration on $\mathcal X$. We continue this process and stop once $\mathrm{index}_p(x)\in \{1,2,...,k\}$ and 
    $$\sum_{i=1}^k p(y^{(k)}_i)>1-p(y^{(k)}_{\mathrm{index}_p(x)}).$$This happens eventually because the partial sums converge to 1 and $p(y^{(k)}_{\mathrm{index}_p(x)})$ is a non-decreasing sequence that is eventually positive. The reason $p(y^{(k)}_{\mathrm{index}_p(x)})$ is eventually positive is because if $p(y^{(k)}_{\mathrm{index}_p(x)})=0$ for all $k$, then the support of $p$ has at most $\mathrm{index}_p(x)-1$ elements, a contradiction. 
    
   Suppose the process stops at $k=n$. When the process stops, we have
   $$
   \{y\in\mathcal X: p(y)\geq p(y^{(n)}_{\mathrm{index}_p(x)})\}=\{y\in\{x_1,..., x_n\}: p(y)\geq p(y^{(n)}_{\mathrm{index}_p(x)})\}, 
   $$
   because all elements outside of $\{x_1,...,x_n\}$ must have probability less than $p(y_{\mathrm{index}_p(x)})$ by the stopping criterion. It follows that $y^{(n)}_{\mathrm{index}_p(x)}=x$. 
\end{proof}

\end{document}